# Human-Like Gaze Behavior in Social Robots: A Deep Learning Approach Integrating Human and Non-Human Stimuli

Faezeh Vahedi, Morteza Memari, Ramtin Tabatabaei, and Alireza Taheri

**Abstract— Nonverbal behaviors, particularly gaze direction, play a crucial role in enhancing effective communication in social interactions. As social robots increasingly participate in these interactions, they must adapt their gaze based on human activities and remain receptive to all cues, whether human-generated or not, to ensure seamless and effective communication. This study aims to increase the similarity between robot and human gaze behavior across various social situations, including both human and non-human stimuli (e.g., conversations, pointing, door openings, and object drops). A key innovation in this study, is the investigation of gaze responses to non-human stimuli, a critical yet underexplored area in prior research. These scenarios, were simulated in the Unity software as a 3D animation and a 360-degree real-world video. Data on gaze directions from 41 participants were collected via virtual reality (VR) glasses. Preprocessed data, trained two neural networks—LSTM and Transformer—to build predictive models based on individuals' gaze patterns. In the animated scenario, the LSTM and Transformer models achieved prediction accuracies of 67.6% and 70.4%, respectively; In the real-world scenario, the LSTM and Transformer models achieved accuracies of 72% and 71.6%, respectively. Despite the gaze pattern differences among individuals, our models outperform existing approaches in accuracy while uniquely considering non-human stimuli, offering a significant advantage over previous literature. Furthermore, deployed on the NAO robot, the system was evaluated by 275 participants via a comprehensive questionnaire, with results demonstrating high satisfaction during interactions. This work advances social robotics by enabling robots to dynamically mimic human gaze behavior in complex social contexts.**

## I. INTRODUCTION

NON-VERBAL behaviors, such as gaze direction and gestures, are foundational in human communication, conveying mental states, intentions and complementing speech [1, 2]. Gaze direction, a crucial part of social interactions [3-5], serves as a key indicator of interests, intentions, and preferences [6-8], surpassing other nonverbal cues in psychological significance [9]. As social robots increasingly assume roles as assistants and companions [10-12], in education, healthcare and collaborative tasks, replicating human-like nonverbal behaviors—especially adaptive gaze—is essential for natural communication [3, 11, 13-18]. However, existing research predominantly focuses on verbal communication, leaving a gap in multiparty interaction contexts where robots must dynamically respond to both human and non-human stimuli.

Prior work has explored robot gaze behavior in storytelling [19], turn taking in a multiparty setting [20], enhancing human participation [21], and human-robot collaboration [22-24]. Aligning closely with our study, Domingo et al. [25, 26] and Lathuilière et al. [14], utilized neural networks for robot gaze control during human-robot interaction in multiparty settings but they overlooked non-human stimuli, a limitation shared by our earlier work [27].

In our previous study [27], we developed an empirical time-motion model of human gaze behavior using deep neural networks (DNNs), achieving 65% accuracy in predicting gaze direction toward humans. However, this model excluded non-human stimuli, a critical limitation given their prevalence in everyday social settings. For instance, environmental cues like doorbells or alarms often redirect human attention, yet robots lacking awareness of these stimuli risk appearing socially inept or disengaged.

To address this gap, we propose a novel empirical motion-time pattern for social robot gaze behavior that integrates both human and non-human stimuli. Our approach advances the field in three key ways:

1. *Incorporation of non-human stimuli* (e.g., door movements, object interactions) enabling robots to dynamically prioritize attention in complex environments.

2. *Designing short and diverse scenarios* to include various realistic social situations, minimizing gaze aversion due to boredom in data collecting process.

3. *Collecting data from a larger group of participants* to enrich our dataset allowing for more generalizable and reliable conclusions.

To achieve this and develop a behavior for social robots that is both natural and mirrors human behavior, it was essential to track human gaze direction in various social situations. Two scenarios including human and non-human stimuli were implemented in Unity: one as a 3D-animation and the other as a 360-degree filmed video of these situations enacted in a real environment. Participants without any

This Work was supported by the "Sharif University of Technology" under Grant No. G4030507. *(Corresponding author: Alireza Taheri).*

Faezeh Vahedi, Morteza Memari, Ramtin Tabatabaei, and Alireza Taheri are with the Social and Cognitive Robotics Laboratory, Center of Excellence in Design, Robotics, and Automation (CEDRA), Sharif University of Technology, Tehran, Iran.

Informed consent was obtained from all individual participants included in the study.

abnormalities affecting gaze direction and vision were invited to view these scenarios using the Meta Quest 2 VR headset virtual reality headset. The gaze direction data of these participants were recorded frame-by-frame by sensors embedded in the VR glasses. Following this, the data was preprocessed to eliminate any incomplete or erroneous entries; and to ensure a balanced representation of different classes, data augmentation techniques were applied. To analyze and extract the gaze direction patterns, two deep neural network models for time-series data were utilized: Long Short-Term Memory (LSTM) and Transformer. The models were evaluated using K-Fold cross-validation and top-k accuracy metrics; and finally, the results were discussed and analyzed and implemented on the NAO robot. Fig. 1 illustrates an overview of this study.

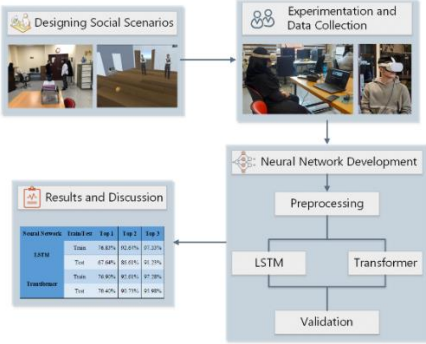

**Fig. 1.** An overview of this study

## II. METHODOLOGY

As it is mentioned earlier, the objective of this study is to develop an empirical model of gaze behavior across different social situations. To collect relevant data, two scenarios were designed to observe and record participants' gaze directions in various social interactions. The first part of the methodology details these scenarios, describing the chosen social contexts and how they were implemented in the study. The second section outlines the process for testing participants and gathering gaze data. Finally, the third part of the methodology focuses on the development of neural networks to analyze this data, covering the preprocessing steps, the design of LSTM and Transformer models, and the methods used to evaluate model performance.

### A. Scenarios of Social Situations

This study aims to develop a motion-time model of human gaze behavior, capturing both where and when people look in social settings. The objective is to enable social robots to predict gaze direction during interactions that involve both human and environmental elements. To collect realistic data, we designed two immersive scenarios, experienced by participants through a virtual reality (VR) headset. These scenarios were created using two approaches: a 3D animation that simulates social situations and a 360-degree video that presents a fully immersive, real-world environment.

### The First Scenario: 3D Animation

In this scenario, various social situations were designed to be implemented as a 3D animation. Building on our previous research [27], which featured only human stimuli, we incorporated several non-human elements to create a scenario that reflects everyday social interactions. As illustrated in TABLE I, this scenario includes both human and non-human stimuli. Human stimuli encompass three characters exhibiting various common social behaviors. All characters were designed identically to mitigate the influence of facial and personality traits on gaze behavior.

TABLE I
THE FIRST SCENARIO STIMULI

| Human stimuli | Non-Human stimuli |
|---|---|
| Standing (silent/speaking) | Footstep sounds |
| Moving (right/left) | Television turning on showing news |
| Moving (straight ahead) | Door opening and closing |
| Waving (silent/speaking) | Television turning on showing statics |
| Standing with arms crossed (silent/speaking) | Phone ringing |
| Engaging in conversation | Ball falling and rolling |
| Entering/exiting the scene | Doorbell ringing |
| Pointing at another character | Alarm clock sound |

The designed scenario comprises 48 permutations of human and non-human stimuli, with each stimulus lasting roughly 5 seconds, totaling approximately 4 minutes. Each scene features 2 to 3 characters positioned at angles between -90 and +90 degrees, with 0 degrees indicating the viewer is looking straight ahead. Characters are placed at varying distances from the viewer to mimic real social situations. Non-human stimuli are also placed at different angles and occur simultaneously with human stimuli to evaluate the prioritization of non-human versus human stimuli. Due to the absence of an eye tracker sensor in the used virtual reality headset (details provided in later sections), human and non-human factors were strategically placed in the scene to require the participant to turn their head to view them.

In this section, we implemented the designed scenario as a 3D animation in the Unity software to make it viewable using virtual reality glasses. Fig. 2 displays four selected frames from the video created in Unity.

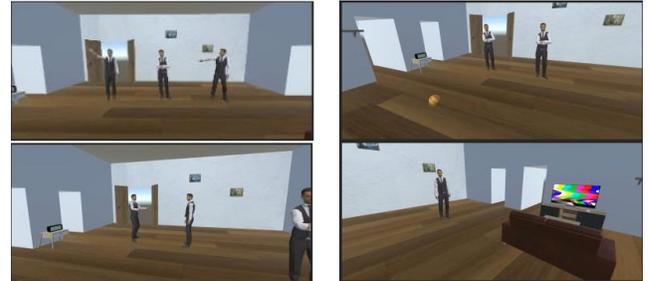

**Fig. 2.** Some frames of the designed 3D animation.

### Second Scenario: 360-degree Video Filming Method

The objective of designing the second scenario was to create more realistic social situations. We aimed to replicate the same social scenarios and include both human and non-human stimuli as in the previous scenario, but with a more natural sequence of events. As in the initial scenario, non-human stimuli were positioned at various angles and appeared simultaneously with human stimuli to evaluate the priority given to each.

Due to the absence of an eye tracker sensor in the used virtual reality headset, human and non-human stimuli were placed in parts of the scene that required the participant to turn their neck to view them. Four individuals (two women and two men) were selected as human agents to perform various social behaviors, including: entering and exiting the room, walking, having conversations, waving, and pointing at objects. Non-human stimuli included an object falling, the screen turning on in static and news modes, a door opening and closing, door knocking sounds, mobile phone alerts, and phone ringing.

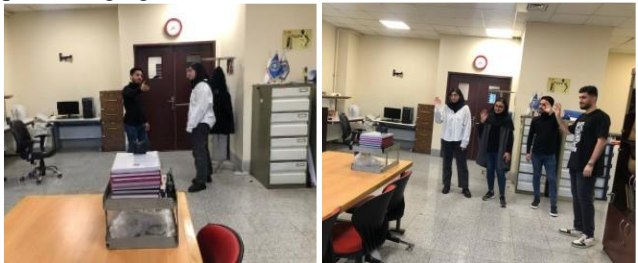

**Fig. 3.** Enacting the second scenario and recording the video using 360-degree camera.

The scenario was implemented and carried out in a real environment rather than being animated. To conduct the tests using VR glasses, it was necessary to record the scenario. For this purpose, the Nikon KeyMission 360 camera, introduced by Nikon in September 2016, was employed. This camera allows for the recording of 360-degree videos in 4K resolution at 24 frames per second. The scenario was enacted by two females and two males and recorded in the Social and Cognitive Robotics Laboratory, Sharif University of Technology, Iran. This recording was then transformed into a nearly two-minute 360-degree video. Fig. 3 displays some selected images from the scenario's execution. To enable playback of this video using virtual reality glasses and to ultimately collect data, the video was integrated into the Unity environment. It was then prepared for VR playback and experimentation.

### B. Experimentation and Data Collection

To test the scenarios and collect the data, the Meta Quest 2 VR headset was used. The Meta Quest 2, released in October 2020 by Reality Labs, a division of Meta Platforms, features a positional tracking system called Oculus Insight. This system uses internal cameras and sensors on the headset to track the participant's movements in physical space. One of these sensors is the IMU (Inertial Measurement Unit), which measures the Quest's movements, such as rotation and acceleration. The Quest 2 provides precise and reliable tracking in the virtual environment based on rotation but lacks an eye tracker sensor and does not track pupil movements.

As previously mentioned, the inability to track pupil position was a serious limitation of this research. Therefore, the scenario was designed in a way to ensure that participants would need to turn their necks to prioritize their gaze direction. Fig. 4 show the use of this device by our participants during the data collection. The data collection took place at the Social and Cognitive Robotics Laboratory, Sharif University of Technology, Iran; with a total of 41 participants (21 women and 20 men) aged 19 to 23 years. The participants had no specific conditions affecting their field of vision or gaze direction, ensuring that the collected data represents the general population. Data collection at each time step included information on the head rotation angles around the x, y, and z axes, as shown in Fig. 4. However, the yaw rotation angle around the y-axis is the primary focus of this research.

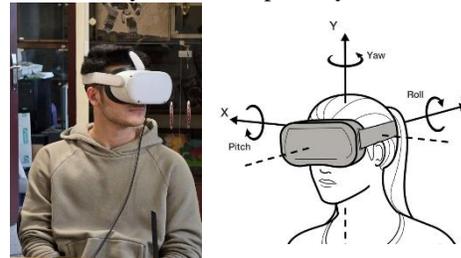

**Fig. 4.** Data collecting setup and Head rotation angles and axes in the Meta Quest 2 VR headset [28].

### C. Neural Network Development

This research aims to design a network to enhance a robot's ability to engage in natural and effective human-robot interactions. Specifically, the goal is to develop a model that can appropriately predict where a social robot should look based on the sequence of social situations that have recently occurred. To develop such a gaze prediction model, two different neural network architectures were used: Long Short-Term Memory (LSTM) and Transformer. These architectures are well-suited for processing and predicting time-series data. The following sections provide a detailed explanation of each step highlighting the preprocessing techniques, the architecture of neural network models, and the validation process.

### Preprocessing

To predict where the robot should look to mimic human behavior, the data recorded by the VR glasses was converted into a format with a time step of 0.1 seconds, ensuring consistent time intervals across all participants. For the LSTM and Transformer networks' input, the data required assigning labels. For both scenarios, each data point was labeled to indicate the part of the image the participant was looking at, based on the recorded angles. Considering participants' gaze range and field of view, the observed angles were segmented into appropriate intervals, resulting in six distinct classes for classification in the first scenario, and seven classes in the second one.

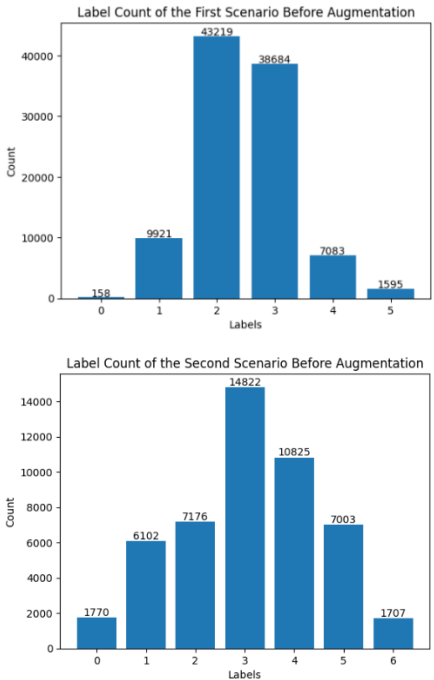

**Fig. 5.** Distribution of the data before data augmentation for each scenario

Fig. 5 illustrates the non-uniform distribution of the data, which adversely affects the learning process of neural networks.

It is noteworthy that the data is fed into the neural network with a sequence length of 30 frames. This means that the time series spans three seconds (30 frames × 0.1 seconds per frame) before decision making, and the gaze direction is determined based on the sequence of events during the last three seconds.

The input to neural networks, regardless of their architecture, should include scene features at each time step along with the data. Therefore, based on the scenario events detailed in previous sections, for each scenario, a scene properties matrix was created. This matrix consists of 25 columns, which are fed into the neural network at each time step. The first column represents the time step, while the remaining 24 columns correspond to features related to each character and object. Essentially, each row of this matrix indicates the human and non-human stimuli present in a time step.

For the first scenario, these stimuli include six human features for each person, totaling 18 elements and 6 non-human stimuli. The second scenario includes 4 features for each person (a total of 16 elements) and 8 non-human stimuli.

*LSTM*

To implement the LSTM models, we used the Keras library in Python. The LSTM (Long Short-Term Memory) is a type of recurrent neural network capable of learning and retaining long-term dependencies, making it suitable for solving time series problems. The first layer is a Bidirectional LSTM layer with 32 neurons, using the tanh activation function. The Bidirectional wrapper allows the LSTM to learn the sequence of events in both forward and backward directions. The second layer is another Bidirectional LSTM layer with 32 neurons and the tanh activation function. This layer includes L1 & L2 regularizers with factors of 0.00001 for L1 and 0.0001 for L2 to prevent overfitting. The next layer is a 20% Dropout layer, which enhances the network's performance by reducing overfitting. The flatten layer converts the inputs into one-dimensional data and passes it to the first dense layer, which has 32 neurons and uses the sigmoid activation function. Finally, the output layer is a dense layer with 6 and 7 neurons for the first and second scenario, respectively, using the softmax activation function to output the probability of belonging to each class. Fig. 6 shows a representation of this architecture. The first scenario's LSTM model has 41702 trainable parameters while the second scenario's model has 41735 trainable parameters.

*Transformer*

The Transformer model is known for its ability to handle long sequential data efficiently, making it suitable for various time series processing tasks. This model was implemented using the Keras library in Python. The model features 2 encoders, each comprising 4 heads and 1024 neurons with swish (or SiLU) activation function. These encoders are arranged sequentially, allowing the model to capture complex patterns in the data. The output vector from the encoders passes through a Maxpool layer, which helps in reducing the dimensionality and focusing on the most significant features.

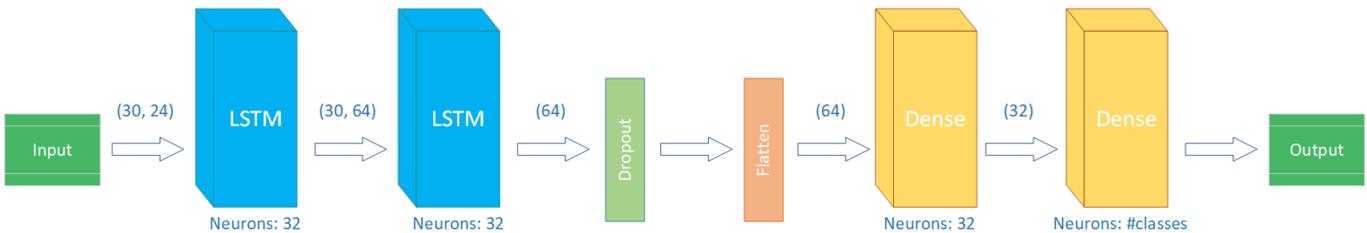

**Fig. 6.** Architecture of the LSTM model.

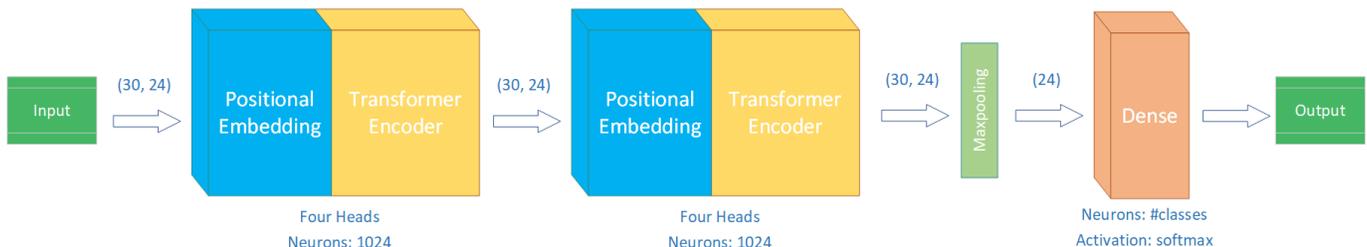

**Fig. 7.** Architecture of the Transformer model.

The final output layer is a dense layer with 6 and 7 neurons for the first and second scenario respectively, utilizing the softmax activation function to determine the probability of belonging to each class. The architecture of this network is illustrated in Fig. 7. The first scenario's Transformer model has 121238 trainable parameters while the second scenario's model has 121263 trainable parameters.

*Validation*

To ensure the robustness and reliability of both architectures, a 10-fold cross-validation technique was employed. The dataset was partitioned into 10 subsets, with each model being trained on 9 subsets and evaluated on the remaining subset in each iteration. This process was repeated 10 times, ensuring that each subset served as the test set once, while the remaining subsets were used for the training. Consequently, each model was trained 10 times, utilizing 90% of the data for training and 10% for testing. The average accuracy across these 10 iterations was reported for each model.

Both architectures utilized the categorical cross-entropy loss function and the Adam optimizer with learning rate of 0.001. Training was conducted for a maximum of 100 epochs. Early stopping with a patience of 10 was applied, meaning training would stop if the validation accuracy did not improve after 10 epochs. The model accuracy was reported using the top-k accuracy score with k = 3.

*D. Implementation on Robot*

To evaluate the models' performances in a more practical manner, we implemented them on a Nao robot and asked participants to fill out a questionnaire about various aspects of robot's social interactions.

A social scenario was enacted by two individuals in the presence of the robot. This scenario included common social situations with human stimuli such as individuals entering and exiting a room, moving, talking, pointing and non-human stimuli including objects falling, a TV turning on and phone ring (similar to scenarios discussed in *A. Scenarios of Social Situations*).

In order to accurately determine the position, sound direction, and assessing robot's surrounding human and non-human stimuli, a Kinect 2 sensor was deployed. The technology of this sensor, allows for detecting individuals and landmarks identification. By utilizing specified thresholds and kinematics energy, the actions and gestures of each individual can be effectively monitored, detecting human stimuli and also moving objects. Additionally, the high-quality microphone arrays of Kinect 2 enable precise detection of sound direction, therefore helping robot to identify non-human stimuli. The

Kinect's microphone array employs advanced beamforming techniques to localize sound sources by analyzing audio signals from multiple microphones. This allows the system to determine the angle of incoming sounds and effectively distinguish between human and non-human cues based on their spatial characteristics. By triangulating the sound source's position relative to the viewer and utilizing algorithms that filter out background noise, the Kinect can accurately identify which stimuli are present in its environment. By combining the features of Kinect 2 with our neural network architectures, our robot demonstrates enhanced comprehension and responsiveness towards human and non-human stimuli in its surroundings.

TABLE II
EVALUATION STATEMENTS

| | |
|---|---|
| **Statement 1:** | I feel satisfied with the interaction of the robot in the video. |
| **Statement 2:** | I'm satisfied with the robot's coordination with people's movements in the video. |
| **Statement 3:** | In my opinion, the robot demonstrates a good understanding of the people in the video. |
| **Statement 4:** | I believe the robot in the video is a great social companion. |
| **Statement 5:** | The robot's interaction with the people in the video felt remarkably human-like to me. |
| **Statement 6:** | The robot in the video appears so lifelike that I can easily imagine it as a living being. |
| **Statement 7:** | The robot in the video pays good attention to the people around it. |
| **Statement 8:** | The robot behaved intelligently in the video. |
| **Statement 9:** | The robot in the video responded adeptly to non-human stimuli. |
| **Statement 10:** | The robot in the video exhibited a solid understanding of its surroundings. |

We recorded the robot's gaze behavior and interaction with this scenario in a one-minute video. This video was then shared with participants to seek their feedback on the robot's behavior by filling out a questionnaire. This questionnaire employed a 5-point Likert scale and included a series of statements inspired by the UTAUT [29] and a survey from a previous study with similar objectives [30] aimed at assessing the performance of the top model trained on the robot. Participants were requested to indicate their level of agreement with ten statements outlined in TABLE II. A 5-point Likert scale was utilized, with verbal descriptors ranging from 'strongly disagree' to 'strongly agree'.

III. RESULTS AND DISCUSSION

In this section firstly, neural network accuracies and then performance evaluation of models is discussed.

### A. Neural Network Accuracies

This section presents and discusses the results obtained from each scenario in the study. The results are divided into three main parts. Firstly, we discuss the accuracy of the models derived from the first scenario on both the training and test data using LSTM and Transformer models. Secondly, the results of the second scenario's models are discussed. Finally, a comprehensive analysis and review of all the results are provided.

#### Results of the First Scenario

As mentioned, K-Fold cross-validation and top-k accuracy scores were employed to validate the results of the LSTM and Transformer models. Fig. 8 depicts accuracy percentages.

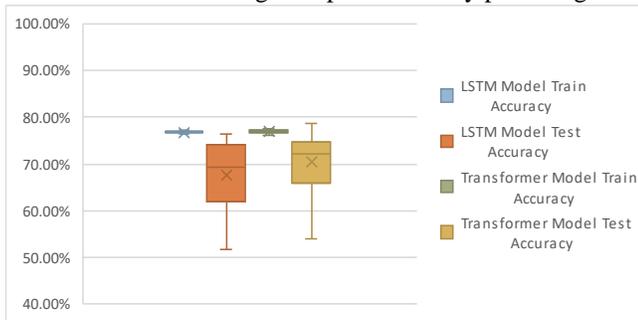

**Fig. 8.** First scenario's models' accuracies.

The average accuracy on the training data is 76.83% and 76.90% for the LSTM and transformer models respectively. Lower standard deviation shows consistency in accuracy values across different folds and thus suggesting that the model is well-designed and has effectively learned from the training data, despite the varying partitions applied by the cross-validation technique. The average accuracy on the test data for the LSTM and Transformer models is 67.64% and 70.40% respectively. The variations in accuracy across different subsets (K) can be attributed to several factors, one of the primary reasons being the differences in individuals' perspectives on visual stimuli, which will be discussed further.

Fig. 9 presents a chart depicting the recorded data on the direction of gaze of individuals in the first scenario. The horizontal axis represents time, while the vertical axis represents the angle of head rotation. The blue lines indicate the average angles of head rotation, and the gray shaded area represents the standard deviation of these angles.

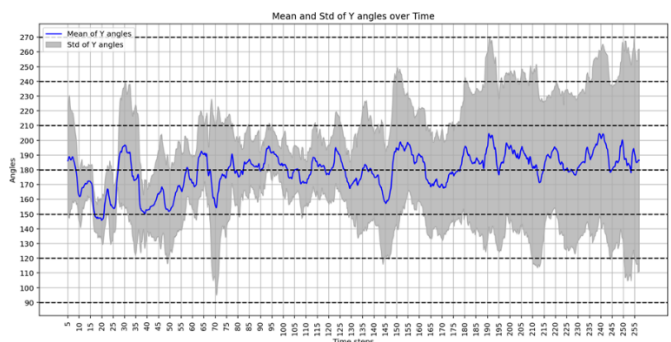

**Fig. 9.** Mean and Standard Deviation chart of the first

scenario's data.

In this scenario, the angles are defined such that 180 degrees indicates looking straight ahead. Turning the head to the right increases the angle, reaching 270 degrees at the farthest right position. Conversely, turning the head to the left decreases the angle from 180 degrees, reaching 90 degrees at the farthest left position. Human stimuli were primarily placed in the center of the image, covering the range of 160 to 200 degrees, while non-human stimuli were mostly located in the corners of the image.

It is observed that the average data fluctuates around 180 degrees with a ±30-degree variation. This phenomenon can be explained in two ways:

- Given the details of the first scenario, individuals looked less at the corners of the image where non-human stimuli were located, indicating a preference for human stimuli over non-human stimuli.
- Instead of turning their heads, individuals moved their pupils to look at the non-human stimuli in the corners of the image. However, due to the lack of access to an eye tracker sensor, this data was not captured.

Another noteworthy observation is the large gray area, particularly during certain time intervals, which indicates a high standard deviation. This phenomenon suggests that individuals differ in their decision-making regarding the direction of gaze, and that various visual stimuli hold different priorities in their minds. It can also be observed that the standard deviation increases over time; this may be attributed to participants' gaze aversion due to boredom.

Despite the mentioned issues, the LSTM and Transformer models predict the gaze direction well, with average accuracies of about 68% and 70%, respectively. TABLE III provides a summary of the results of the two neural network models for the first scenario. The Table shows that the Transformer model consistently outperformed the LSTM model on the test data. Notably, when we provided the models with an opportunity to reconsider their gaze direction, the accuracy percentages improved noticeably. Specifically, by allowing the model to consider the top three output probabilities rather than just the highest probability, we achieved remarkable results. The LSTM model achieved an accuracy of 91.23%, while the Transformer model excelled with an impressive 95.98% accuracy.

We processed the input data corresponding to the first scenario and obtained the model's output to visualize the results of our neural network models. This output was then imported into the Unity software, where we synchronized the camera movement with the generated data. Fig. 10 displays some frames extracted from the resulting video. The center of the blue ellipse represents the average angles of the individuals' gaze direction. The ellipse's longest diameter corresponds to the standard deviation of the recorded data from the participants in the experiment. In some frames, the gaze direction of individuals is highly dispersed, resulting in a larger diameter of the ellipse. Despite this, the model has accurately predicted the gaze direction and as shown in Fig. 10, human and non-human stimuli are correctly followed by the robot's gaze direction.

TABLE III
MODEL ACCURACY RESULTS FOR THE FIRST SCENARIO

| Neural Network | Train/Test | Top 1 | Top 2 | Top 3 |
|----------------|------------|-------|-------|-------|
| LSTM | Train | 76.83% | 92.64% | 97.33% |
| | Test | 67.64% | 86.61% | 91.23% |
| Transformer | Train | 76.90% | 92.61% | 97.28% |
| | Test | 70.40% | 90.73% | 95.98% |

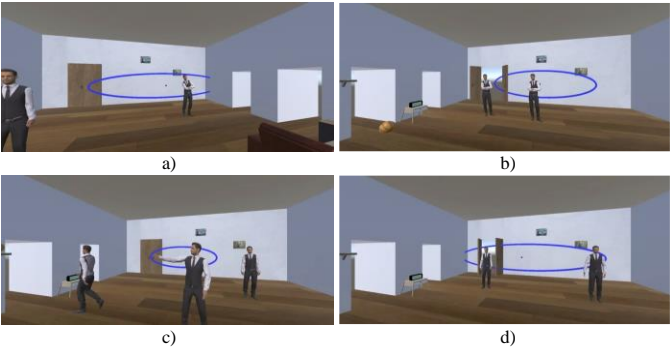

**Fig. 10.** Frames from the video implemented using the neural network model outputs for the first scenario; a) the person on the right is speaking; b) the door is closing; c) the person in the middle is pointing to the person exiting; and a) a ball is falling.

*Results of the Second Scenario*

Similar to the first scenario, K-Fold cross-validation and top-k accuracy scores have been utilized for result validation of the LSTM and Transformer models. The accuracy percentages of models are depicted in Fig. 11.

The average accuracy on the training data is 74.85% and 74.48% for LSTM and Transformer models, respectively. Notably, the accuracy values on this dataset exhibit minimal fluctuations. This phenomenon suggests that the model, despite being evaluated on different data partitions using the cross-validation technique, is well-designed and effectively captures the training data. The mean accuracy on the test dataset is 72.04% for LSTM and 71.59% for Transformer model. The observed variations in accuracy across different subsets (K) can be attributed to several factors, with one of the primary reasons being the differences in individuals' interpretations of visual stimuli, which will be elaborated upon further.

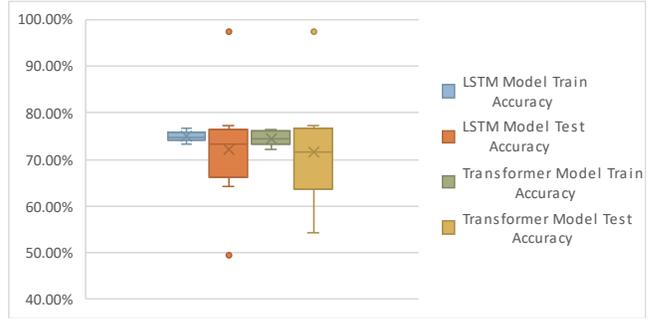

**Fig. 11.** Second scenario's LSTM model accuracies.

Fig. 12 illustrates a chart representing the recorded gaze direction data of individuals in the second scenario. The horizontal axis denotes time, while the vertical axis indicates the angle of head rotation. The blue lines represent the mean angles of head rotation, and the gray shaded area signifies the standard deviation of these angles.

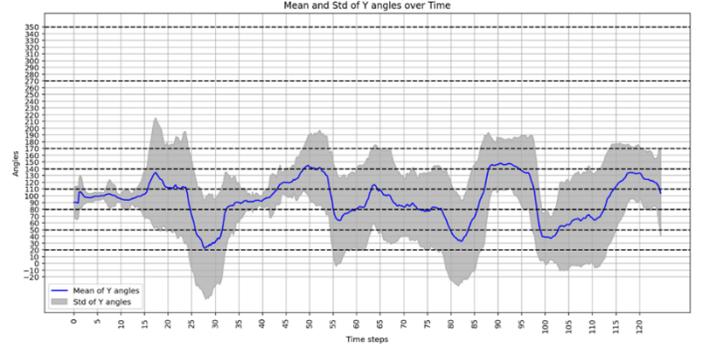

**Fig. 12.** Mean and Standard Deviation chart of the second scenario's data.

In this scenario, the angles are defined as follows: a 90-degree angle represents looking straight ahead. Turning the head to the right increases the angle, reaching 180 degrees at the maximum rightward position of the neck (without involving body rotation). Conversely, turning the head to the left decreases the angle from 90 degrees, reaching 0 degrees at the maximum leftward position.

The large gray area, particularly during certain time intervals, indicates a high standard deviation from the mean. This suggests significant variability in individuals' gaze direction decisions, highlighting that different visual stimuli are prioritized differently by individuals.

Despite the mentioned issues, the models predict the angle class with an accuracy of approximately 72%. TABLE IV provides a summary of the results for the two neural network models in the second scenario. It is observed that, in general, both models yield nearly identical results. Furthermore, providing the models with a second chance significantly enhances their accuracy. Specifically, allowing the robot to select the gaze direction based on the top three output probabilities, rather than the highest probability alone, increases the models' accuracy on the test dataset to 86.83% for the LSTM model and 88.76% for the Transformer model which are very promising results.

TABLE IV

MODEL ACCURACY RESULTS FOR THE SECOND SCENARIO

| Neural Network | Train/Test | Top 1 | Top 2 | Top 3 |
|---|---|---|---|---|
| LSTM | Train | 74.85% | 87.70% | 93.58% |
| | Test | 72.04% | 80.73% | 86.83% |
| Transformer | Train | 74.48% | 87.27% | 93.44% |
| | Test | 71.59% | 83.41% | 88.76% |

Finally, to visualize the results of the neural network models, the input data corresponding to the video of the second scenario was fed to the models and the predicted gaze behavior were obtained from the models' output. This output was then imported into the Unity software, where the camera movement was synchronized with the generated data. Fig. 13 presents frames extracted from the resulting video.

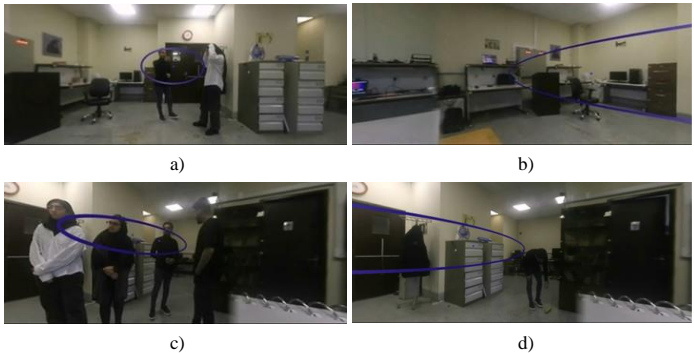

a)          b)

c)          d)

**Fig. 13.** Some selected frames from the video implemented using the neural network model outputs for the second scenario; a) the stimulus is a person's gesture; b) the screen turning on; c) a conversation between individuals; and d) an object falling to the ground.

The center of the blue ellipse represents the average gaze angles of the individuals and the longest diameter indicates the standard deviation of the recorded data from the participants in the experiment. It is observed that the gaze direction of individuals shows significant dispersion in some frames, resulting in a larger ellipse diameter. Despite this, the model has accurately predicted the gaze direction and the robot correctly follows human and non-human stimuli.

*Summary of Results*

In this section, the results of the deep neural networks, LSTM and Transformer, were thoroughly examined and analyzed for each scenario. Additionally, the neural network outputs were visualized in the Unity environment to gain a better understanding of the extracted gaze patterns. Below is a brief summary of the results for both scenarios.

TABLE V
RESULTS OF THE MODELS' ACCURACIES FOR TEST DATA

| Neural Network | The First Scenario | The Second Scenario |
|---|---|---|
| LSTM | 67.64 % | **72.04 %** |
| Transformer | **70.40 %** | 71.59 % |

As illustrated in TABLE V, the Transformer model demonstrated superior performance in the first scenario, achieving an accuracy of 70.40% on the test data, which is approximately 3% higher than the LSTM model. In the second scenario, the performance difference between the LSTM and Transformer models was less pronounced, with the LSTM model achieving a higher accuracy of 72.04% on the test data.

Previous literature has predominantly reported the results of gaze pattern models qualitatively, often excluding non-human stimuli from consideration. Consequently, precise comparisons are challenging. However, compared to the study by Tabatabaei [27], which achieved 65% accuracy on their data by focusing solely on human stimuli, this research demonstrates superior performance with an accuracy of 72%. Additionally, this study incorporates non-human factors and stimuli in extracting gaze patterns, representing an advancement over previous literature which often overlooked the impact of non-human stimuli on gaze direction.

It is important to note that individuals' gaze direction in various social situations is not uniform and depends on multiple factors [31]. People differ in their decision-making regarding gaze direction, and various visual stimuli hold different priorities for them. Consequently, individuals' gaze patterns vary, making the extraction of these patterns by neural networks highly challenging. If the accuracy on the training data increases beyond a certain threshold, the model risks overfitting, which can lead to a significant drop in accuracy on the test data.

*B. Performance Evaluation*

As discussed in previous sections, a video showcasing a robot utilizing our best model to interact and respond to social situations was recorded. This video, along with a questionnaire, was shared as a Google Form, to which 275 individuals responded in a 48-hour span.

The mean and standard deviation of questionnaire results are presented in TABLE VI. For all statements, the mean score is above the midpoint, indicating participants strongly agree with the statements, and consequently the desirable behavior of the robot in response to environmental stimuli. The averages of 4.21 and 4.20 for statements 7 and 2, respectively, demonstrate that the robot pays good attention to its surroundings and reacts accordingly, which is satisfactory to the participants. In statement 5, the mean score of 3.84 suggests that the robot's behavior, in terms of gaze direction, is comparable to human behavior. The standard deviation of 1.27 for this statement indicates variability in participants' perspectives, which aligns with results reported in the literature. Statement 6 recorded the lowest mean of 3.63 and the highest standard deviation of 1.30. These statistics suggest that participants did not perceive the robot as closely resembling a living being. Given the restriction of robot's behavior to changing only the gaze direction, this outcome is logical, and it is not expected for the robot to be perceived as a living being merely by mimicking gaze angle.

In conclusion, the results of this evaluation suggest that participants in the survey rated the robot's attentiveness to its surroundings as very high, and they believe the robot has the ability to behave in a manner similar to humans in social situations. It is important to note that individuals' preferences and priorities for appropriate gaze direction may vary in certain scenarios. However, considering the substantial number of participants, it can be concluded that the robot is generally capable of imitating human gaze behavior effectively.

TABLE VI
MEAN AND STANDARD DEVIATION OF QUESTIONNAIRE RESULTS

| Statements | Mean | Standard Deviation |
|---|---|---|
| Statement 1 | 4.13 | 1.12 |
| Statement 2 | 4.20 | 1.12 |
| Statement 3 | 4.18 | 1.14 |
| Statement 4 | 3.81 | 1.27 |
| Statement 5 | 3.84 | 1.27 |
| Statement 6 | 3.63 | 1.30 |
| Statement 7 | 4.21 | 1.10 |
| Statement 8 | 4.15 | 1.17 |
| Statement 9 | 4.17 | 1.09 |
| Statement 10 | 4.00 | 1.19 |

## IV. LIMITATIONS AND FUTURE WORK

Our research encountered several challenges. Notably, the lack of access to virtual reality glasses equipped with eye tracker sensors hindered accurate data recording. Participants sometimes moved their pupils instead of their heads to look at certain stimuli, resulting in unrecorded gazes toward these specific stimuli. Another challenge was the variability and diversity of individuals' gaze behavior in different social situations. Gaze direction is not uniform and depends on various factors. Individuals differ in their decision-making regarding gaze behavior, and different visual stimuli have different priorities in their minds. Therefore, gaze patterns vary among individuals, making pattern extraction by neural networks highly challenging. Future work could involve exploring alternative neural network architectures or regression-based approaches.

## V. CONCLUSIONS

In this study, we explored empirical motion-time patterns of human gaze behavior in various social situations to determine human-like gaze directions for robots following specific events. We designed and implemented two scenarios within the Unity software environment. Forty-one participants viewed these scenarios using virtual reality glasses, enabling frame-by-frame recording of their visual field data. This data was preprocessed to remove incomplete entries, and data augmentation techniques were applied to balance the class distribution. Scene properties matrices were constructed for each scenario, and two types of deep neural networks, LSTM and Transformer, were employed to extract gaze direction patterns. The models were evaluated using K-Fold cross-validation and top-k accuracy metrics. For the first scenario, the LSTM and Transformer models predicted the test data with accuracies of 67.6% and 70.4%, respectively. For the second scenario, the LSTM and Transformer models achieved test data accuracies of 72.0% and 71.6%, respectively. To further assess the models' performance, they were deployed on a Nao robot. A comprehensive questionnaire was administered to 275 participants, revealing high levels of satisfaction during interactions.

Despite the challenges, the model presented in this research outperformed existing approaches in terms of accuracy and considered non-human stimuli, which is a significant advantage over previous literature. Furthermore, utilizing virtual reality glasses equipped with eye tracker sensors in future steps, can result in a more accurate measurement of gaze patterns towards various human and non-human stimuli.

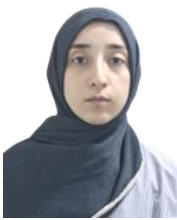

**Faezeh Vahedi** received the bachelor's degree in Mechanical Engineering from Sharif University of Technology, Tehran, Iran in 2024. She is currently studying masters of Mechanical Engineering at University of Tehran. Her research interests include robotics, machine learning, control theory, and mechatronics.

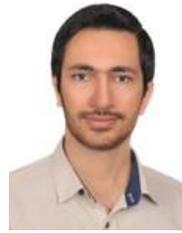

**Morteza Memari** received the B.S. and M.S. degrees in Mechanical Engineering from Sharif University of Technology, Tehran, in 2023. He is now a Ph.D. candidate in Mechanical Engineering in Sharif University of Technology, Iran since Fall 2023. From 2023, he was a Research Assistant with the Social and Cognitive Robotics Laboratory. Also, he was a teaching assistant in more than 5 undergraduate and graduate courses in the field of robotics, AI, and control. His research interests include social robotics, artificial intelligence, human-robot interaction, and control theory. Mr. Memari was a recipient of the best Mechanical Engineering B.S. Project Award in 2021, and he is a member of the Center of Excellence in Design, Robotics, and Automation (CEDRA).

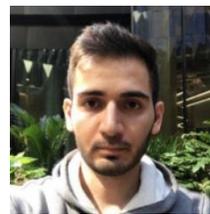

**Ramtin Tabatabaei** received the B.S. and M.S. degrees in Mechanical Engineering from University of Tehran and Sharif University of Technology, respectively. Currently, he is a PhD student at the University of Melbourne, Australia. His research interests include robotics, machine learning, and human-robot interaction.

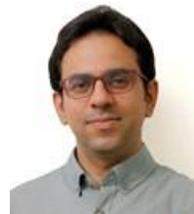

**Alireza Taheri** is an Associate Professor of Mechanical Engineering with an emphasis on Social and Cognitive Robotics at Sharif University of Technology, Tehran, Iran. He is the Head of the Social and Cognitive Robotics Lab. and the Measurement Systems Lab. at Sharif University of Technology. The line of his research focuses on designing/using Social and Cognitive Robotics, Virtual Reality Systems, and Human-Robot Interaction (HRI) platforms for education and rehabilitation of children with special needs (e.g. children with autism, children with hearing problems, children with cerebral palsy). His researches include robots' design and fabrication, serious games' design, artificial intelligence and control, conducting educational/clinical interventions for children, developing cognitive architectures for social robots, mathematical modeling of participants' behaviors during HRI, and empowering robots to analyze users' behaviors automatically and then react adaptively.